%% file: root.tex
\documentclass[letterpaper, 10 pt, conference]{ieeeconf}  

\IEEEoverridecommandlockouts                              

\usepackage{amsmath,amsfonts}
\usepackage{algorithmic}
\usepackage{algorithm}
\usepackage{array}
\usepackage{caption}
\usepackage{textcomp}
\usepackage{stfloats}
\usepackage{url}
\usepackage{verbatim}
\usepackage{graphicx}
\usepackage{cite}
\usepackage{kotex}
\usepackage{bbm}
\usepackage{multirow}
\usepackage{booktabs}
\usepackage{amsmath}
\usepackage{amssymb}
\usepackage{dsfont}
\usepackage{makecell}

\title{\LARGE \bf
Pixel2Catch: Multi-Agent Sim-to-Real Transfer for Agile Manipulation with a Single RGB Camera
}

\author{Seongyong Kim, Junhyeon Cho, Kang-Won Lee, and Soo-Chul Lim\\
Dongguk University
\thanks{}}

\begin{document}

\twocolumn[{%
\renewcommand\twocolumn[1][]{#1}%
\maketitle
\begin{center}
    \vspace{-5mm}
    \centering
    \captionsetup{type=figure}
    \includegraphics[width=0.97\linewidth]{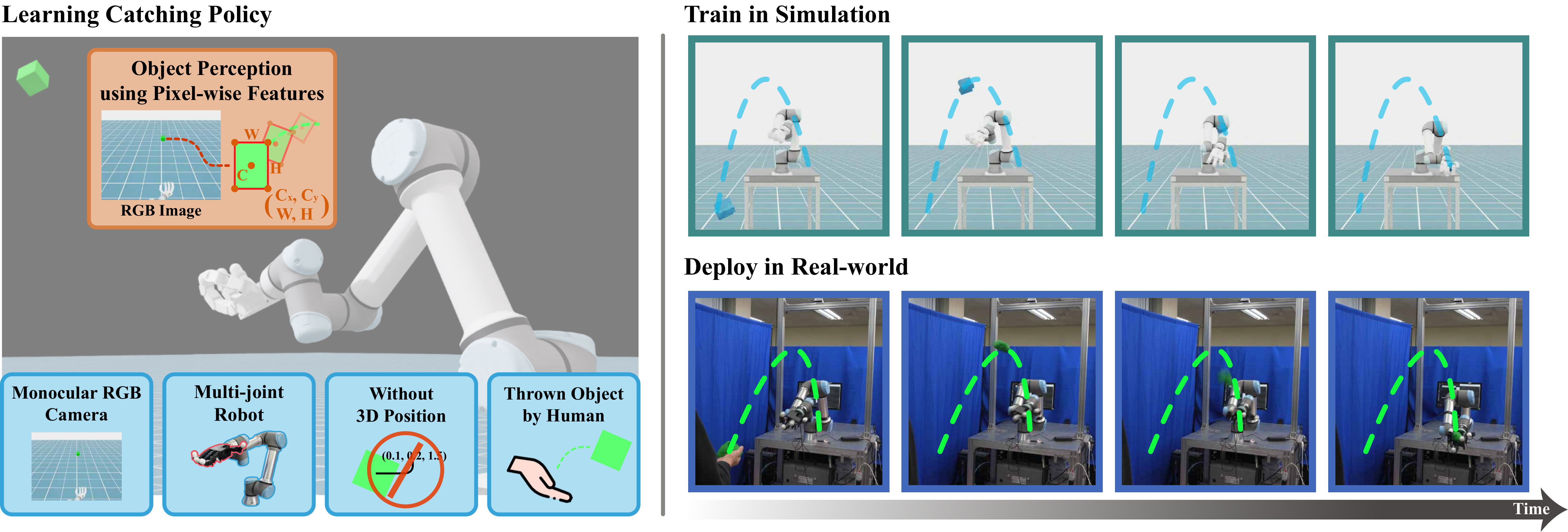}
    \captionof{figure}{
    We propose \textbf{Pixel2Catch}, an RGB-only robotic catching system without explicit 3D position estimation.
    The system consists of a robot arm equipped with a multi-fingered hand and a single RGB camera.
    Inspired by human visual perception, object motion is inferred from pixel-level features in image space rather than metric 3D coordinates.
    Policies trained in simulation are transferred directly to the real robot without fine-tuning.
    }
    \label{fig:abstract}
\end{center}
}]
\vspace{-5mm}

\thispagestyle{empty}
\pagestyle{empty}

\vspace{-3mm}
\begin{abstract}
To catch a thrown object, a robot must be able to perceive the object’s motion and generate control actions in a timely manner.
Rather than explicitly estimating the object’s 3D position, this work focuses on a novel approach that recognizes object motion using pixel-level visual information extracted from a single RGB image.
Such visual cues capture changes in the object’s position and scale, allowing the policy to reason about the object’s motion.
Furthermore, to achieve stable learning in a high-DoF system composed of a robot arm equipped with a multi-fingered hand, we design a heterogeneous multi-agent reinforcement learning framework that defines the arm and hand as independent agents with distinct roles.
Each agent is trained cooperatively using role-specific observations and rewards, and the learned policies are successfully transferred from simulation to the real world.

\end{abstract}

\section{Introduction}
\input{tex/1_intro}

\section{Related Work}
\input{tex/2_related_works}

\section{Robot System for Catching Thrown Objects}
\input{tex/3_system_setup}

\section{Learning Catching Policies}
\input{tex/4_method}

\section{Experimental Results}
\input{tex/5_results}

\section{Conclusion and Future Work}
\input{tex/6_conclusion}

\bibliographystyle{IEEEtran}
\bibliography{reference}

\end{document}

%% file: tex/1_intro.tex
When humans catch a thrown object, they do not rely on explicit 3D position values or precise metric representations of the object.
Instead, they perceive the object’s motion by observing relative changes in its apparent position and size over time, and generate catching movements based on these visual cues.
This highlights that dynamic catching can be achieved through the interpretation of visual motion patterns, rather than explicit estimation of the object’s 3D state.

However, most prior studies on dynamic object manipulation have relied on the motion capture systems~\cite{kim2014catching,chen2024cushioncatch,salehian2016dynamical,tassi2025ima,lan2023dexcatch} or depth sensors~\cite{zhang2024catch,huang2023dynamic,chen2021deep,kim2025learning} to estimate the object’s 3D position.
While such sensing modalities enable explicit geometric measurements, the performance of the resulting control models is strongly influenced by sensor quality.
In particular, accurately estimating 3D object positions in real-world environments is often more challenging than in simulation, which leads to a significant sim-to-real gap when transferring learned policies.

Motivated by principles of human visual perception, this work proposes a dynamic catching framework that focuses on recognizing object motion through pixel-level visual cues observed in a single RGB image, rather than explicitly estimating the object’s 3D position.
To extract such visual cues, we employ Segment Anything Model 2 (SAM 2)~\cite{ravi2024sam}, which provides robust object segmentation and enables the extraction of pixel-level information capturing relative changes in image space position and scale.
By leveraging these relative visual changes over time, the proposed approach represents object motion without relying on precise metric measurements, making it less sensitive to sensor accuracy.

In addition, to enable effective learning of grasping behaviors, we employ a multi-agent reinforcement learning (MARL)~\cite{yu2022mappo} framework.
This design allows efficient control of high-DoF robotic systems, such as a robot arm equipped with a multi-fingered hand.
While previous MARL studies~\cite{li2025real,liu2025language,ding2020multi_arm,zhan2024multi_hand} have primarily focused on cooperation among multiple robots with homogeneous structures, this work decomposes a single robotic system into heterogeneous agents by modeling the robot arm and the robot hand as independent agents.
Through this heterogeneous MARL formulation, the arm and hand can be trained cooperatively with role-specific objectives, enabling stable and coordinated dynamic grasping.

The proposed framework is first trained in a simulation environment that is designed to closely mirror the real robotic system.
This enables the policies learned in simulation to be directly deployed on the physical robot without additional fine-tuning.
In real-world experiments, the robot successfully catches objects thrown by a human using the learned policies.
Furthermore, we evaluate the effectiveness of the proposed approach by comparing it against policies trained without pixel-level visual cues and those learned using a single-agent formulation.

In summary, the main contributions of this paper are as follows:
\begin{itemize}
\item We propose a novel framework for dynamic dexterous manipulation that represents object motion using pixel-level features extracted from a single RGB camera.
\item We adopt a single-stage heterogeneous MARL framework by decomposing a high-DoF robotic system into a robot arm and a multi-fingered hand, each trained with role-specific observations and rewards.
\item Through system identification and domain randomization, we successfully transfer policies trained in simulation to the real world, demonstrating agile and stable catching of human-thrown objects using only RGB visual input.
\end{itemize}
\vspace{2mm}

%% file: tex/2_related_works.tex
\subsection{Perception and Manipulation for Dynamic Objects}
Robotic manipulation research has been studied across a wide range of scenarios, from static manipulation tasks in which the object is already held by the robot~\cite{yin2023rotating,wang2024lessons} or fixed at a specific location~\cite{lee2024dextouch,lin2024learning,lee2025progressive}, to dynamic manipulation tasks involving freely moving objects~\cite{wang2025spikepingpong,dastider2024unified}.
While static manipulation assumes that the object state does not change significantly over time, dynamic manipulation requires the ability to perceive and respond to continuous changes in the object’s position and state.

Many prior studies on dynamic manipulation have focused on scenarios in which object motion is confined to a 2D plane.
Examples include grasping objects moving along a conveyor belt~\cite{yamamoto2024real, chen2021deep}, objects following circular trajectories in a two-dimensional plane~\cite{fuentes2009binocular}, or objects rolling on a flat surface~\cite{lee2024catching}.
In these scenarios, object motion follows a relatively constrained pattern within the observation space, making the perception and control problems more tractable.

To handle more complex scenarios, recent studies have extended dynamic manipulation to objects moving freely in 3D space, typically relying on position estimation methods to recognize object motion.
One representative approach leverages RGB-D cameras.
Huang et al.~\cite{huang2023dynamic} and Zhang et al.~\cite{zhang2024catch} estimated the 3D coordinates of objects using RGB-D cameras and implemented policies for catching thrown objects in real-world environments.
These methods segment the object using color-based segmentation, extract the corresponding depth values, and compute 3D coordinates using the camera’s intrinsic and extrinsic parameters.
However, in real-world settings, accurate position estimation is often more challenging than in simulation, as it strongly depends on sensor reliability, which can lead to performance degradation.

Another approach employs marker-based motion capture systems~\cite{yu2021neural, yan2024impact, nguyen2025dipp}.
Chen et al.~\cite{chen2024cushioncatch} and Tassi et al.~\cite{tassi2025ima} used such systems to collect human demonstrations and object trajectory data, which were then used to train control models.
While this approach enables accurate position estimation, it requires attaching markers to the object and incurs high system setup costs, limiting its practicality in real-world applications.

Overall, existing dynamic manipulation studies tend to rely on explicit position estimation or additional sensing modalities.
In contrast, inspired by human visual perception, this work adopts an alternative approach that recognizes object motion using pixel-level visual information directly observed from a single RGB image, without explicitly estimating the object’s 3D position.
This approach simplifies the input representation, reduces sensor dependency, and enables effective dynamic catching of thrown objects.

\subsection{Learning Frameworks for Catching Policy}
Previous studies on dynamic object catching have commonly employed a robotic arm equipped with a simple end-effector, such as basket or scoop-shaped designs, to train policies~\cite{chen2024cushioncatch, abeyruwan2023agile, gold2022catching, dastider2024retro, tassi2025ima, dong2020catch}.
In these approaches, the catching process is determined by the approach trajectory of the robot arm, while the role of the end-effector is limited to passively receiving the object.
As a result, the control problem is simplified around arm-level motion, allowing  catching policies to be learned using a single control model.

To encourage more human-like catching motions and to stably handle objects with diverse shapes, prior studies have explored systems equipped with multi-fingered hands~\cite{lan2023dexcatch, kim2025learning, hu2023modular}.
For example, Zhang et al.~\cite{zhang2024catch} proposed a two-stage reinforcement learning framework in which the catching skill is learned in a sequential manner using a multi-fingered hand.
While this approach achieves stable performance through staged learning, it typically requires stage-wise reward design and policy transfer between stages, leading to increased complexity in the training pipeline.

In this work, we adopt a MARL approach as an alternative to such multi-stage learning schemes, aiming to learn a catching policy within a single training stage.
Most existing MARL studies focus on cooperation among multiple robotic systems with homogeneous structures, such as UAV swarm control~\cite{tang2025dnn}, autonomous driving~\cite{liu2025language}, bimanual or multi-hand manipulation~\cite{ding2020multi_arm, zhan2024multi_hand}, and cooperative control of multiple quadruped robots~\cite{feng2024multi_quadruped}.

In contrast, our approach focuses on decomposing a single robotic system into heterogeneous components based on their functional roles, rather than coordinating multiple robots.
Specifically, we model the robot arm and the robot hand as independent agents and design role-specific observations and reward functions for each agent.
This formulation allows the arm agent to concentrate on motion control for approaching the object, while the hand agent focuses on forming stable grasps during the catching process.
As a result, effective catching policies can be learned within a single training process, without requiring staged learning or policy transfer.

\begin{figure*}
\centering
\includegraphics[width=0.8\textwidth]{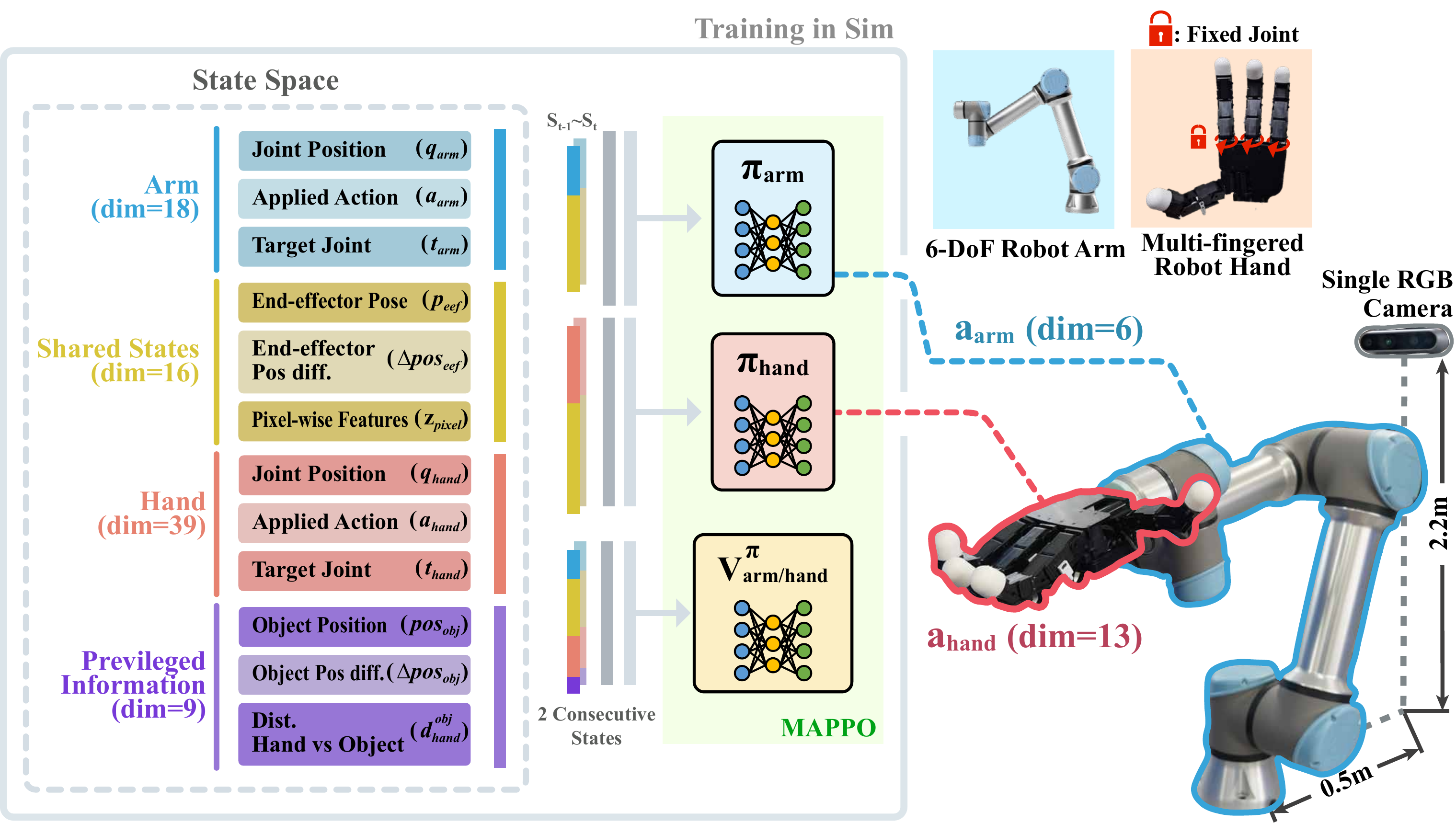}
\caption{Pipeline of the system and experimental setup.
Each policy ($\pi_{arm}, \pi_{hand}$) operates on selected observations from two consecutive timesteps.
Privileged information is used only during value network training.
A single RGB camera is mounted 0.5\,m behind and 2.2\,m above the robot.
The arm and hand are controlled by separate policies that are trained collaboratively to catch a thrown object.}
\label{fig:pipeline}
\vspace{-5mm}
\end{figure*}

%% file: tex/3_system_setup.tex
\subsection{Task Description}
The goal of this work is to control a robot arm and hand to stably catch a thrown object without dropping it.
To achieve this, we decompose the catching task into two complementary roles: the arm positions the hand in a region suitable for catching, while the hand focuses on securely grasping the object.

In real-world environments, unlike in simulation, obtaining precise 3D coordinates of a thrown object is challenging.
Rather than relying on direct position measurements, we infer object motion from visual cues.
Inspired by human perception, which interprets object motion through changes in size and position within the field of view, we extract pixel-level features from a single RGB image.
By leveraging these visual features, our approach enables effective learning of catching policies without requiring explicit 3D object positions.

\subsection{Real-world Setup}
We constructed an experiment using a high-DoF robot system and a camera for catching thrown objects as follows:

\begin{itemize}
\item \textbf{Arm.}
The Universal Robots UR5e, a 6-DoF robot arm, is utilized in this work.
The robot is mounted on a table at a height of 0.81 m, and its initial posture is configured to ensure that the hand is oriented toward the expected trajectory of the object.
\end{itemize}

\begin{itemize}
\item \textbf{Hand.}
The Allegro Hand is attached to the robot arm, with three joints fixed to simplify control and maintain grasp stability, as shown in Fig.~\ref{fig:pipeline}.
As a result, 13 joints are controlled using a target joint position controller.
\end{itemize}

\begin{itemize}
\item \textbf{Camera.}
A single RealSense D435 camera is fixed in the environment to provide a global view of the object's overall motion.
To mimic human visual perception, our system is designed to use only RGB images, without depth information.
\end{itemize}

\begin{figure}[t]
\centering
\includegraphics[width=0.97\linewidth]{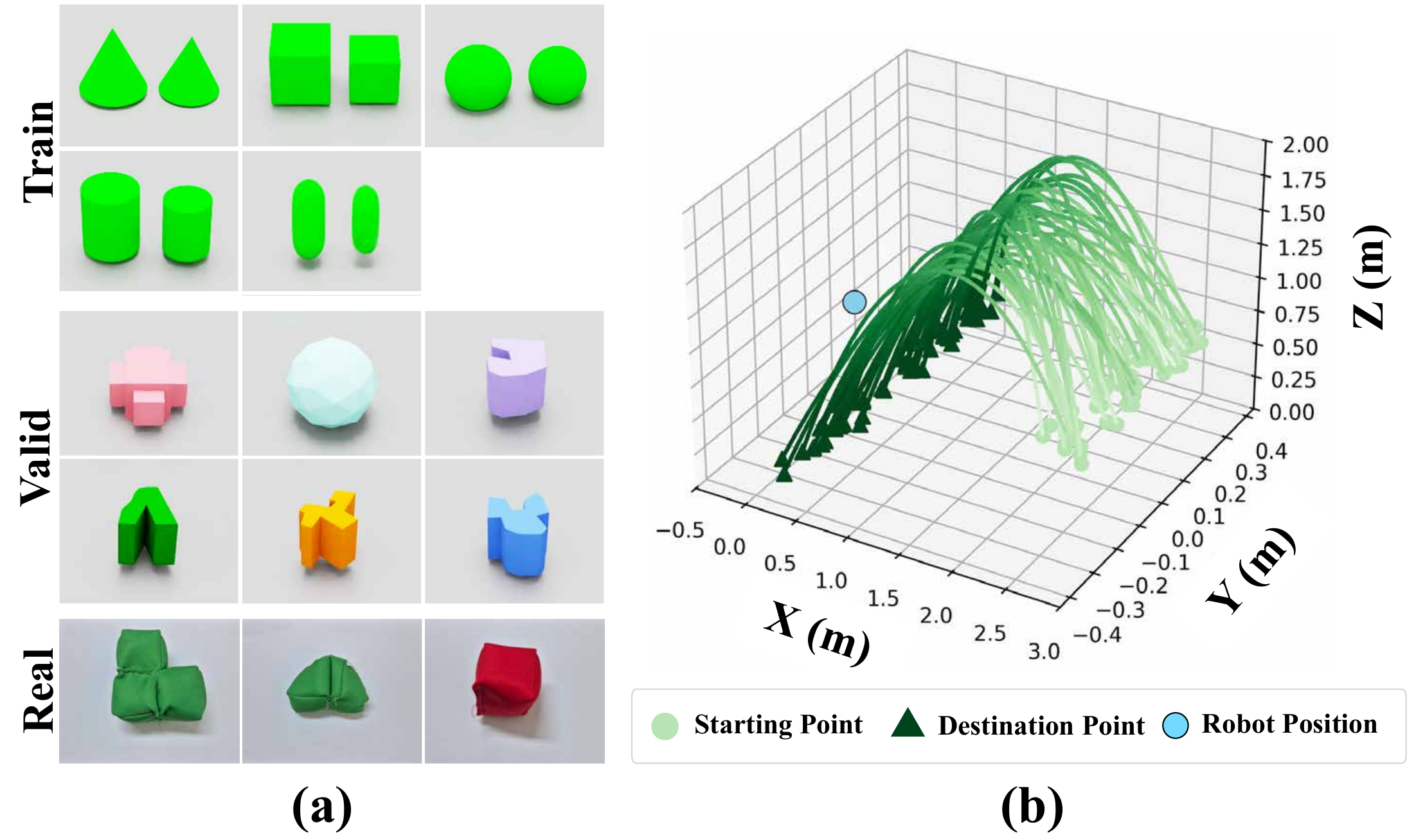}
\caption{(a) Objects used for training (top), validation (middle) in simulation, and real-world experiments (bottom).
(b) Random object trajectories generated in simulation, shown without robot motion to highlight object dynamics.}
\label{fig:object_trajectories}
\vspace{-5mm}
\end{figure}

\subsection{Simulation Setup}
We develop our simulation environment using the NVIDIA Isaac Lab framework \cite{mittal2025isaaclab} to closely match the real robot system.
The simulated robot employs a USD model identical to the real hardware, and the camera is modeled as a pinhole camera with the same field of view as the Intel RealSense D435 used in real-world experiments.
To improve robustness to variations in object motion, we randomize the object’s initial state at the beginning of each episode, including its position, orientation, linear velocity, and mass (Fig.~\ref{fig:object_trajectories}(b)).
During training, five objects with distinct geometries are used, and their scales are varied to encourage generalization across different shapes and sizes (Fig.~\ref{fig:object_trajectories}(a)).
To align the simulation control loop with the real robot, we apply a control decimation of 4, resulting in a 30~Hz control policy operating on a 120~Hz physics simulation by repeating each selected action over four consecutive simulation steps.
\vspace{1mm}

%% file: tex/4_method.tex
\subsection{Problem Formulation}
We formulate the catching task as a Multi-Agent Markov Decision Process (MAMDP).
Control is decomposed between an arm policy $\pi_{arm}$ responsible for positioning and a hand policy $\pi_{hand}$ focused on grasping.
Each policy $\pi_{i} ( i \in \{\text{arm}, \text{hand}\})$ receives its own observation $s_t^i$ and selects an action $a_t^i$ to maximize its expected discounted return $E[\sum_{k=0}^T \gamma^k r_k^i]$, where $r_k^i$ is the reward assigned to policy $i$ at timestep $k$.
An episode terminates if the object is dropped or the maximum timestep $T$ is reached.
We utilize Multi-Agent Proximal Policy Optimization (MAPPO)~\cite{yu2022mappo} to train the policies $\pi_{arm}$ and $\pi_{hand}$.

\subsection{Pixel-level Features from a single RGB image}
\begin{figure}
\centering
\includegraphics[width=0.97\linewidth]{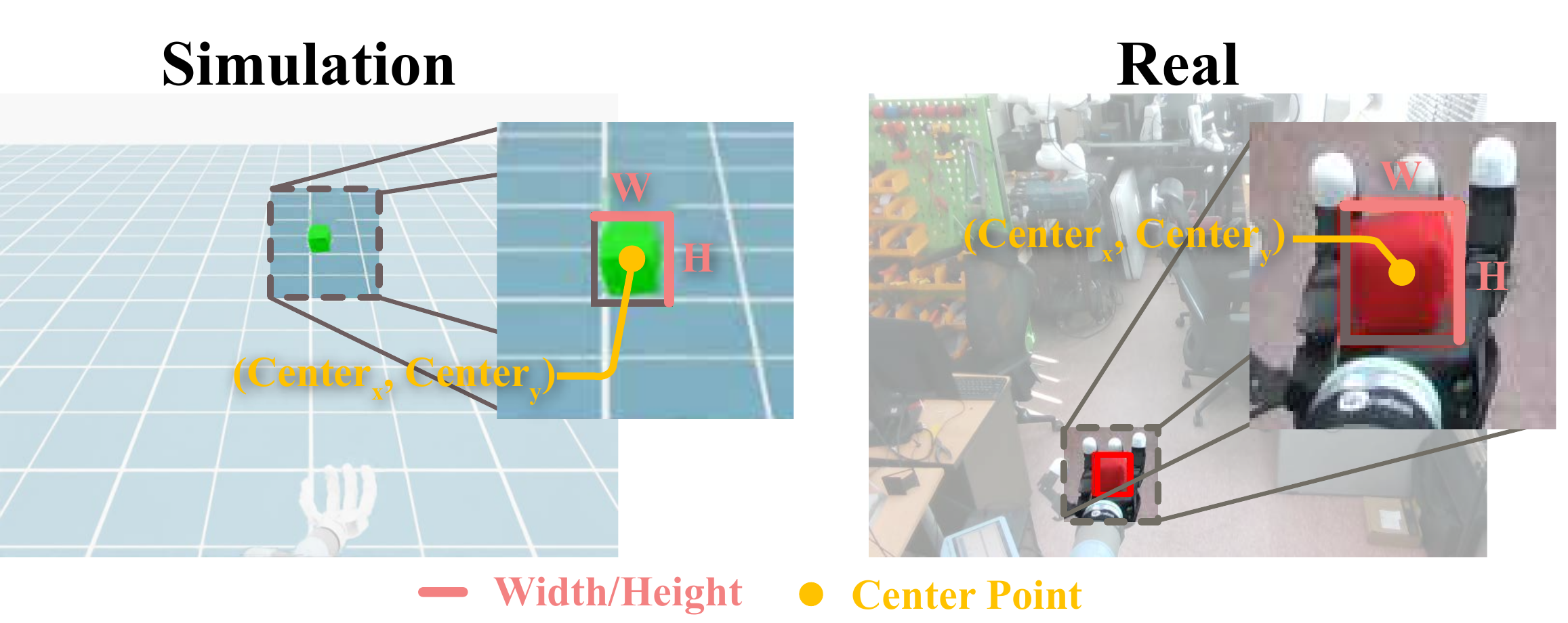}
\caption{Visualization of pixel-level features in simulation and real-world environments. A bounding box is generated around the object in the RGB image, from which the corner and center coordinates, as well as width and height, are extracted. The final input features include these values and their temporal differences.}
\label{fig:pf}
\vspace{-5mm}
\end{figure}

Previous studies have commonly relied on additional sensing modalities to explicitly estimate the 3D position of objects.
In contrast, we defined pixel-level features \(z_{pixel}\) extracted directly from RGB images to recognize object motion.
\(z_{pixel}\) is constructed from visual cues computed from the image space corresponding to the target object.
As illustrated in Fig.~\ref{fig:pf}, we extract the object’s image space center coordinates, width, and height.
The horizontal and vertical components of the center represent the object’s relative lateral and vertical position, while the width and height encode scale changes that implicitly reflect variations in the distance between the robot and the object.
Since pixel-level features from a single timestep are insufficient to capture object motion, we incorporate temporal differences between consecutive timesteps into \(z_{pixel}\).
This allows the policy to infer motion dynamics directly from relative visual changes.

To improve robustness under real-world conditions, we introduce random perturbations of up to 5 pixels to the corner coordinates of the object region during training.
This augmentation accounts for segmentation uncertainty caused by real-world visual variations.

The resulting pixel-based feature vector is defined as:
\begin{equation}
    \label{pixel_features}
       z_{pixel} \in \mathbb{R}^{6} = \{c_{x}, c_{y}, \Delta c_{x}, \Delta c_{y}, \Delta w, \Delta h\}
\end{equation}

Unlike approaches that employ high-dimensional latent representations from image encoders such as ResNet~\cite{he2016deep}, our method directly utilizes task-relevant visual quantities that can be extracted from raw images.
This design choice reduces representation mismatch between simulation and real-world environments and improves learning efficiency by focusing on control-oriented visual information rather than generic semantic features.

Finally, to address the sensitivity of color-based masking methods~\cite{huang2023dynamic,zhang2024catch} to lighting changes and background complexity, we employ SAM2~\cite{ravi2024sam} for object segmentation.
SAM2 enables robust and consistent extraction of object regions across diverse lighting conditions and backgrounds, thereby providing stable pixel-level inputs for real-world deployment.

\subsection{State and Action Space}
As shown in Fig.~\ref{fig:pipeline}, each agent’s state includes pixel-level features \(z_{pixel}\in \mathbb{R}^{6}\), end-effector pose information \(pose_{eef} \in \mathbb{R}^{7}\), and agent-specific joint states (\(q_{arm} \in \mathbb{R}^{6}\), \(q_{hand} \in \mathbb{R}^{13}\)) and actions (\(a_{arm} \in \mathbb{R}^{6}\), \(a_{hand} \in \mathbb{R}^{13}\)), concatenated over two consecutive timesteps.

The object's position \(p_{object} \in \mathbb{R}^{3}\) is not included in the policy input but is utilized solely for training the value network to enhance the generalization and stability of learning.
To provide temporal context, each state space concatenates sequential observations over 2 consecutive timesteps.
For the value network, the input additionally includes observations from both agents together with the object's position, aggregated over 2 timesteps.

Each agent's policy network outputs an action that is used to control its corresponding robot system.
For the arm, the action \(a_{arm}\), generated by the arm policy \(\pi_{arm}\), is added to the current joint positions and executed through a PD controller.
For the hand, the action \(a_{hand}\), produced by the hand policy \(\pi_{hand}\), is rescaled according to the joint limits and used as the target joint position.

\subsection{Reward Design for Multi-agent RL}
To effectively train role-specific agents, we design separate reward functions for the arm and hand policies.
The arm policy \(\pi_{arm}\) is trained to control the robot arm, focusing on approaching the thrown object with the end-effector.
In contrast, the hand policy \(\pi_{hand}\) is trained to control the robot hand to stably grasp the object without dropping it.
The reward functions for the arm and the hand are formulated as follows:
\begin{equation} 
\label{arm_reward}
\begin{split}
        R_t^{\text{arm}} =& r_{time} + r_{\text{dist}}^{\text{palm}} + \lambda_{succ} \mathds{1}_{succ} + \lambda_{app}\mathds{1}_{app} \\ &- \lambda_{fail} (\mathds{1}_{drop} + \mathds{1}_{coll}) - \lambda_{act} ||a_t^{\text{arm}}||^2
\end{split}
\end{equation}

\begin{equation}
    \label{hand_reward}
    \begin{split}
        R_t^{\text{hand}} =& \frac{1}{5} ( r_{\text{dist}}^{\text{palm}} + \sum_{i \in \mathcal{F}} r_{\text{dist}}^{i}) + \lambda_{succ} \mathds{1}_{succ} \\& - \lambda_{fail} (\mathds{1}_{drop} + \mathds{1}_{coll}) - \lambda_{act} ||a_t^{\text{hand}}||^2 
    \end{split}
\end{equation}
where $r_{\text{dist}}$ is defined as the temporal difference in Euclidean distance $d(\cdot, \cdot)$ between the robot link and the object:
\begin{equation}
    \begin{split}
&r_{\text{dist}}^{k} = d(p_{t-1}^{k}, p_{t-1}^{\text{obj}}) - d(p_{t}^{k}, p_{t}^{\text{obj}}),\\& k\in \{palm, thumb,index,middle,ring\}
    \end{split}
\end{equation}
Here, $p_{t}^{\text{obj}}$ denotes the object position, and $k$ represents the link position, which is either the palm or a fingertip.

The terms $\mathds{1}_{succ}$, $\mathds{1}_{drop}$, $\mathds{1}_{app}$, and $\mathds{1}_{coll}$ are binary indicators for a successful catch, object drop, approach, and collision, respectively.
An action penalty term $||a_{t}||^2$ is employed to prevent jerky motions and ensure control stability.
The coefficients are set to $\lambda_{succ}=10.0$, $\lambda_{fail}=5.0$, $\lambda_{app}=0.1$, $\lambda_{act}=0.01$, and $r_{time}=-0.01$.

\subsection{Training Procedure}
We train the policies using the skrl library~\cite{serrano2023skrl} based on the MAPPO framework.
While standard MAPPO implementations typically employ shared policy and value networks across agents, our setup requires separate networks due to the heterogeneity of the agents, which differ in their degrees of freedom and functional roles in the catching task.
The training follows the Centralized Training with Decentralized Execution (CTDE), where a centralized value function is used during training, while each agent executes its own policy independently at test time.
Both the policy and value networks for each agent consist of three fully connected layers with hidden dimensions of [512, 256, 128], and ELU~\cite{nair2010rectified} is used as the activation function.
The model is trained using the following hyperparameters: a clipping parameter of $\epsilon = 0.2$, a discount factor of $\gamma = 0.99$, and a KL-divergence threshold of 0.016.
Training is conducted across 512 parallel environments distributed over two NVIDIA RTX A6000 GPUs.

\subsection{System Identification and Domain Randomization}
To reduce the sim-to-real gap arising from discrepancies in robot dynamics, we perform system identification prior to policy training.
A diverse set of joint-space trajectories is executed on the real robot, and the resulting motions are recorded.
Using these data, joint-level dynamic parameters in simulation, including actuation gains, damping coefficients, friction, and armature, are optimized to minimize trajectory tracking errors between simulated and real executions.

In addition, random noise is applied to both observations and actions during training.
All domain randomization settings used in the training phase are summarized
in Table~\ref{tab:domain_randomization}.

\renewcommand{\arraystretch}{1.1}
\begin{table}[t]
\centering
\caption{Domain randomization parameters applied in training.}
\label{tab:domain_randomization}
\resizebox{0.95\columnwidth}{!}{
\begin{tabular}{c l l l l}
\toprule[1pt]
 & Parameter & Type & Distribution & Range \\
\midrule
\multirow{5}{*}{\makecell{Robot\\Arm}}
 & Joint stiffness   & Scaling  & Uniform & $[0.8,\,1.2]$ \\
 & Joint damping     & Scaling  & Uniform & $[0.8,\,1.2]$ \\
 & Action noise      & Additive & Gaussian    & $\mu=0.0,\ \sigma=0.03$ \\
 & Observation noise & Additive & Gaussian    & $\mu=0.0,\ \sigma=0.005$ \\
 & Initial joint pos         & Additive & Uniform     & $[-0.125,\,0.125]$ \\
\midrule
\multirow{4}{*}{\makecell{Robot\\Hand}}
 & Joint stiffness   & Scaling  & Uniform & $[0.7,\,1.3]$ \\
 & Joint damping     & Scaling  & Uniform & $[0.7,\,1.3]$ \\
 & Action noise      & Additive & Gaussian    & $\mu=0.0,\ \sigma=0.02$ \\
 & Observation noise & Additive & Gaussian    & $\mu=0.0,\ \sigma=0.005$ \\
\midrule
\multirow{2}{*}{Object}
 & Mass              & Scaling  & Uniform     & $[0.5,\,1.5]$ \\
 & Restitution       & Set      & Uniform     & $[0.0,\,0.5]$ \\
\bottomrule
\end{tabular}
}
\vspace{-3mm}
\end{table}

%% file: tex/5_results.tex
In this section, we compare the performance of our Pixel2Catch system with several baseline methods in both simulation and real-world environments.
Across all experiments, performance is evaluated using two quantitative metrics designed to assess both reaching and grasping quality.
The tracking rate (T.R.) measures the proportion of trials in which the palm of the robot hand successfully makes contact with the thrown object, reflecting the policy’s ability to control the arm.
The success rate (S.R.) further evaluates whether the object is stably caught without being dropped after contact, assessing the effectiveness of the overall catching behavior.

In all experiments, objects are thrown toward the robot from random initial positions with randomly sampled directions and velocities.
In simulation, each policy is evaluated on both training objects and unseen objects by executing 10 trials in each of 100 parallel environments.
In real-world experiments, performance is assessed over 30 trials for each object.

\subsection{Baselines}
To validate the effectiveness of our proposed framework and the importance of visual features, we compare our method against the following baselines:

\textit{i) w/o-PF (Without Pixel-level Features):} No pixel-level features are used.
The policy receives only the initial object position and the robot’s proprioception.

\textit{ii) S-A RL (Single-Agent Reinforcement Learning):} PPO \cite{schulman2017ppo} is used to train this baseline, where the arm and hand are controlled by a single agent. Observations and rewards are jointly integrated during training.

\textit{iii) Only-Center (Using Only Center Point Information):} This policy is an ablation of the visual cues, where the policy is trained using only the centroid coordinates($c_x, c_y$).

\textit{iv) Only-WH (Using Only Width and Height Information):} In contrast to \textit{Only-Center}, this policy utilizes only the width and height ($w, h$) from the pixel-level features.

\subsection{Contribution of Pixel-level Features}
Fig.~\ref{fig:sim_traincurve} shows the average tracking rate and success rate of the proposed method over the training process.
Table~\ref{valid_result} shows evaluation results in simulation using both seen objects and unseen objects.
The main findings are summarized as follows:

(i) \textit{w/o-PF}, which does not utilize pixel-level features, achieves the lowest tracking and success rates.
As illustrated in Fig.~\ref{fig:object_trajectories}, the thrown objects follow diverse trajectories rather than moving along a straight path from their initial positions.
Under such conditions, relying solely on the object’s initial position is insufficient to accurately catch objects.
These results highlight that continuous information about object motion is essential for reliable tracking and catching motion.

(ii) Ablation studies were conducted to analyze which features are most critical.
\textit{Only-WH}, which encodes object width and height in image space, indirectly reflects the object’s relative distance to the robot.
However, it lacks sufficient information to infer the object’s motion direction, resulting in poor tracking and success rates.
In contrast, both \textit{Only-Center} and the proposed method can estimate the object’s motion direction using the image space center coordinates, leading to substantially improved tracking and success rates under the simulated trajectory distribution.
Notably, the proposed method, which jointly leverages both center coordinates and scale information, consistently outperforms the single-cue variants.
This indicates that combining directional cues from center motion with distance-related cues from scale variation provides a more complete and robust representation of object motion in image space.

\subsection{Comparison of Multi-agent and Single-agent Learning}
\begin{figure}[t]
\centering
\includegraphics[width=1.0\linewidth]{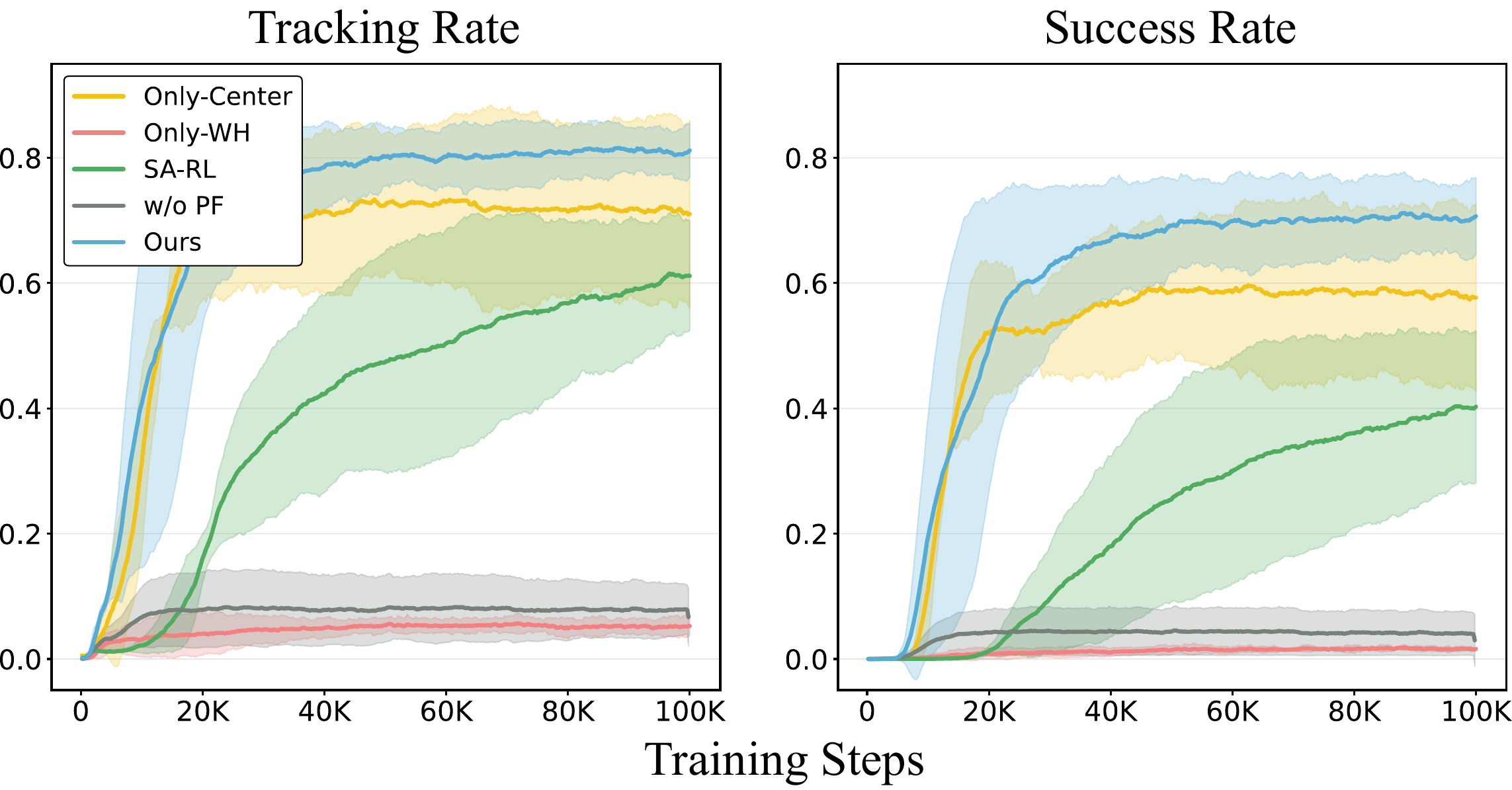}
\vspace{-5mm}
\caption{Tracking and success rates over training. Results are averaged over 3 seeds, and the shaded regions indicate the standard deviation.}
\label{fig:sim_traincurve}
\end{figure}

As shown in Fig.~\ref{fig:sim_traincurve}, the proposed multi-agent framework consistently outperforms \textit{S-A RL} in both tracking and success rates.
This performance gain stems from decomposing the catching task into specialized roles for the arm and hand agents.
Specifically, the arm agent focuses on positioning the end-effector to catch the object, while the hand agent concentrates on forming stable grasps during the catching phase.
This task decomposition offers two key advantages.
First, it allows the use of role-specific reward functions, providing clearer and less conflicting learning signals than the unified reward used in the single-agent formulation.
Second, each agent operates on a role-specific observation space, reducing unnecessary coupling between reaching and grasping objectives.
In contrast, the single-agent model must simultaneously process combined observations and optimize a single policy for both reaching and grasping, which can hinder learning efficiency and stability.

\renewcommand{\arraystretch}{1.1}
\begin{table}
\centering
\caption{Performance of the baselines in simulation. The results are averaged over 3 seeds.}
\label{valid_result}
\resizebox{0.95\columnwidth}{!}{
    \begin{tabular}{ c  l  c  c } 
        \toprule[1pt]
        \multirow{2}{*}{Metric} & \multirow{2}{*}{Method} & \multicolumn{2}{c}{Objects} \\ \cmidrule(lr){3-4}
        & & Seen Objects & Unseen Objects \\ 
        \midrule
        
        \multirow{5}{*}{T.R. (\%)} 
        & w/o-PF        & 12.13 $\,{\scriptstyle \pm\,1.27}$ & 12.11 $\,{\scriptstyle \pm\,1.9}$ \\
        & only-WH        & 8.03 $\,{\scriptstyle \pm\,0.6}$ & 6.11 $\,{\scriptstyle \pm\,0.35}$ \\
        & only-Center        & 87.07 $\,{\scriptstyle \pm\,1.62}$ & 87.5 $\,{\scriptstyle \pm\,1.53}$ \\
        & S-A RL       & 78.2 $\,{\scriptstyle \pm\,1.32}$ & 75.83 $\,{\scriptstyle \pm\,2.00}$ \\
        & Proposed     & \textbf{89.97}$\,{\scriptstyle \pm\,0.21}$ & \textbf{89.28} $\,{\scriptstyle \pm\,0.79}$ \\ 
        \midrule
        
        \multirow{5}{*}{S.R. (\%)} 
        & w/o-PF        & 8.93 $\,{\scriptstyle \pm\,1.04}$ & 7.89 $\,{\scriptstyle \pm\,1.46}$ \\
        & only-WH        & 5.53 $\,{\scriptstyle \pm\,0.51}$ & 4.00 $\,{\scriptstyle \pm\,0.17}$ \\
        & only-Center        & 81.27 $\,{\scriptstyle \pm\,0.76}$ & 80.72 $\,{\scriptstyle \pm\,0.38}$ \\
        & S-A RL       & 63.5 $\,{\scriptstyle \pm\,0.85}$ & 65.44 $\,{\scriptstyle \pm\,2.25}$ \\
        & Proposed     & \textbf{84.13} $\,{\scriptstyle \pm\,0.50}$ & \textbf{84.83} $\,{\scriptstyle \pm\,1.17}$ \\ 
        \bottomrule[1pt]
    \end{tabular}
}
\vspace{-2mm}
\end{table}

\begin{figure*}
\centering
\includegraphics[width=0.9\linewidth]{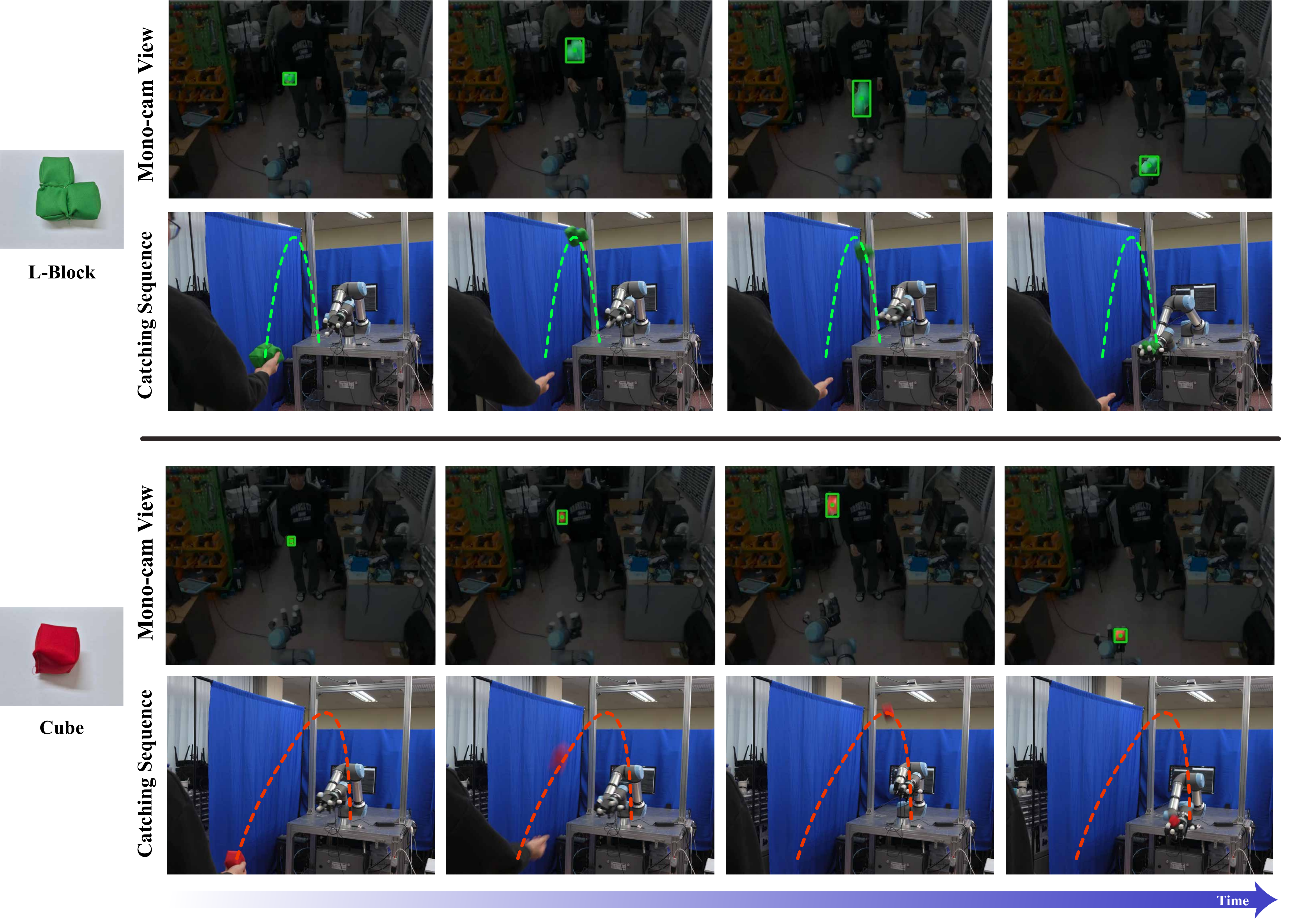}
\caption{Visualization of the results of deploying a trained policy in real-world experiments. This figure presents real-world catching sequences for objects with different geometries (see Fig.~\ref{fig:object_trajectories}(a)).
The ``mono-cam view'' shows the scene captured by the installed camera, where objects are segmented using SAM2, and pixel-level features extracted from the segmentation results are provided to the policy as input.} 
\label{fig:real_valid}
\vspace{-5mm}
\end{figure*}

\subsection{Sim-to-Real Transfer Results}
We evaluate the performance of policies trained in simulation by deploying them on a real robot.
The real robot system operates under ROS2 for communication, and target joint position commands generated by the policies are sent to the arm and hand controllers at 30~Hz.
The task involves catching objects thrown by a human, and representative real-world catching sequences are shown in Fig.~\ref{fig:real_valid}.
Importantly, all policies are transferred directly from simulation to the real robot without any additional real-world fine-tuning.

As shown in Table~\ref{tab:real_result}, the proposed method achieves the best overall performance in real-world experiments, demonstrating strong sim-to-real transfer capability.
In contrast, the baseline methods exhibit significant performance degradation when deployed in the real world.

\textit{Only-WH}, which excludes image-space center information, fails to achieve any successful catch in real-world evaluation.
Without center coordinates in the image space, the policy cannot reliably infer object motion direction, resulting in zero tracking and success rates.

\textit{Only-Center} achieves strong performance in simulation but degrades substantially when transferred to the real world.
Although it frequently achieves initial contact with the object, it often fails to maintain stable grasping, leading to oscillatory hand motions and eventual drops.
This limitation arises from the absence of width and height information, which prevents the policy from inferring relative distance and approach speed between the robot and the object.

The \textit{S-A RL} baseline, which controls both the arm and hand using a single policy, achieves a success rate of approximately 24\%.
In real-world settings, out-of-distribution disturbances can corrupt the shared representation, causing errors in arm and hand control simultaneously and resulting in unstable behavior during sim-to-real transfer.
In contrast, the proposed multi-agent formulation decouples arm reaching and hand grasping, making the policy more robust to such disturbances.

Overall, the observed performance degradation of the baseline methods in real-world experiments can be attributed to differences in object trajectory distributions between simulation and the real world.
While simulation randomizes initial object position, velocity, and direction, real-world throws are affected by unmodeled factors such as object deformation, elasticity, and aerodynamic effects, leading to trajectories that deviate from the simulated distribution.
As a result, methods that rely on a single visual cue, such as \textit{only-WH} and \textit{only-Center}, exhibit limited generalization to these out-of-distribution trajectories.

In contrast, the proposed method jointly leverages both center coordinates and scale variation in image space.
This enables implicit inference of object motion and approach dynamics, allowing the policy to maintain robust performance under diverse real-world throwing conditions.
Consequently, the proposed method achieves a tracking rate of approximately 70\% and a catching success rate of around 50\% in real-world experiments.
\vspace{3mm}

\renewcommand{\arraystretch}{1.1}
\begin{table}
\centering
\caption{Performance of the baselines in the real-world. Results are averaged over 30 trials for each object.}
\label{tab:real_result}
\resizebox{0.95\columnwidth}{!}{
    \begin{tabular}{ c  l  c  c  c } 
        \toprule[1pt]
        \multirow{2}{*}{Metric} & \multirow{2}{*}{Method} & \multicolumn{3}{c}{Objects} \\ \cmidrule(lr){3-5}
        & & Cube & L-block & Triangle \\ 
        \midrule
        
        \multirow{4}{*}{T.R. (\%)} 
        & only-WH        & 0  & 0 & 0 \\
        & only-Center        & 57 & 54 & 54 \\
        & S-A RL       & 50 & 43 & 47 \\
        & Proposed     & \textbf{73} & \textbf{73} & \textbf{70} \\
        \midrule
        
        \multirow{4}{*}{S.R. (\%)} 
        & only-WH        & 0 & 0 & 0 \\
        & only-Center        & 13 & 3 & 23 \\
        & S-A RL       & 33 & 20 & 20  \\
        & Proposed     & \textbf{63} & \textbf{43} & \textbf{43} \\ 
        \bottomrule[1pt]
    \end{tabular}
}
\vspace{-3mm}
\end{table}

%% file: tex/6_conclusion.tex
In this work, we proposed a multi-agent reinforcement learning framework for robotic catching that leverages pixel-level visual features from a single RGB camera.
By decomposing the control problem into arm and hand agents with role-specific objectives, our approach achieves robust performance in both simulation and real-world experiments, consistently outperforming baseline methods.
These results demonstrate that effective dynamic catching can be realized without explicit 3D object position estimation, relying solely on relative visual cues from RGB observations.

Despite these promising results, the current study is limited to a single-arm robotic setup.
As future work, we plan to extend the proposed framework to bimanual robotic platforms to enable more stable and robust catching of objects with diverse sizes and geometries.